\documentclass{Interspeech2024}

\interspeechcameraready 

\usepackage[natbib, bibencoding=utf8, citestyle=numeric, bibstyle=ieee, maxbibnames=999, maxcitenames=2, mincitenames=1, sortcites]{biblatex}
\bibliography{mybib}

\usepackage[capitalize]{cleveref}
\usepackage[activate]{microtype}
\usepackage[acronym, shortcuts, nohypertypes={acronym}]{glossaries}
\newacronym{DL}{DL}{deep learning}
\newacronym{DNN}{DNN}{deep neural network}
\newacronym{ISWF}{ISWF}{isoelastic social welfare function}
\newacronym{ML}{ML}{machine learning}
\newacronym{MLP}{MLP}{multi layer perceptron}
\newacronym{SER}{SER}{speech emotion recognition}
\newacronym{UAR}{UAR}{unweighted average recall}

\newcommand{\ie}{{i.\,e.,} }
\newcommand{\eg}{{e.\,g.,} }
\newcommand{\base}{{\textbf{\textsc{Base}}} }
\newcommand{\persn}{{\textbf{\textsc{Pers}}$_N$} }
\newcommand{\perse}{{\textbf{\textsc{Pers}}$_E$} }
\newcommand{\persa}{{\textbf{\textsc{Pers}}$_A$} }

\title{Enrolment-based personalisation for improving \\individual-level fairness in speech emotion recognition}

\name[affiliation={1,2}]{Andreas}{Triantafyllopoulos}
\name[affiliation={1,2,3,4}]{Björn}{Schuller}

\address{
  $^1$CHI -- Chair of Health Informatics, MRI, Technical University of Munich, Germany \\
  $^2$MCML -- Munich Center for Machine Learning $^3$MDSI -- Munich Data Science Institute \\
  $^4$GLAM -- Group on Language, Audio, \& Music, Imperial College London, UK}
\email{andreas.triantafyllopoulos@tum.de}

\keywords{personalisation, fairness, speech emotion recognition, computational paralinguistics, deep learning}

\begin{document}

\maketitle

\begin{abstract}
    
    The expression of emotion is highly individualistic.
    However, contemporary speech emotion recognition (SER) systems typically rely on population-level models that adopt a `one-size-fits-all' approach for predicting emotion.
    Moreover, standard evaluation practices measure performance also on the population level, thus failing to characterise how models work across different speakers.
    In the present contribution, we present a new method for capitalising on individual differences to adapt an SER model to each new speaker using a minimal set of enrolment utterances.
    In addition, we present novel evaluation schemes for measuring fairness across different speakers.
    Our findings show that aggregated evaluation metrics may obfuscate fairness issues on the individual-level, which are uncovered by our evaluation, and that our proposed method can improve performance both in aggregated and disaggregated terms.
\end{abstract}

\section{Introduction}

\Ac{SER} was one of the earliest computational paralinguistic tasks to be tackled by \ac{ML} and remains a core focal point for the speech community~\citep{Schuller18-SER}.
Progress in \ac{SER} research has largely been measured in terms of gains in performance, with recent advances in \ac{DL} spearheading current efforts~\citep{Wagner23-DOT}.
This performance is usually measured in terms of a single metric computed over a held-out test set, typically accuracy or \ac{UAR} for classification or correlation for regression.

However, recent investigations on the \emph{bias} of \ac{ML} methods have called for increased attention to alternative performance evaluations that account for this bias.
Related efforts have been largely geared towards identifying and mitigating bias resulting from certain \emph{group} characteristics, such as biological sex, gender, or ethnicity~\citep{Gorrostieta19-GDB, Wagner23-DOT}.
However, some works are calling attention to the need for an individual-level measure of fairness~\citep{Fan22-IIS, Wagner23-DOT}, \ie quantifying performance separately for each speaker.

This is related to the well-known problem of diverging performance across different speakers in several speech technology tasks (\eg see Doddington's zoo~\citep{Doddington98-SGL}), a topic which is becoming increasingly relevant when considering recent ethical and legal guidelines.
According to the EU AI Act regulations adopted by the European Parliament~\citep[Amendment 52, ``Proposal for a regulation'', Recital 26 c]{AIAct}:
\begin{quote}
    There are serious concerns about the scientific basis of AI systems aiming to identify or infer emotions, particularly as expression of emotions vary considerably across cultures and situations, and even within a single individual.
    % Among the key shortcomings of such systems are the limited reliability, the lack of specificity and the limited generalizability.
    [...]
    Therefore, AI systems identifying or inferring emotions or intentions of natural persons on the basis of their biometric data may lead to discriminatory outcomes and can be intrusive to the rights and freedoms of the concerned persons.
\end{quote}
This quote illustrates the importance that lawmakers place on the generalisability of \ac{SER} systems across different individuals, which in turn requires researchers to ensure equal treatment for all users of an \ac{SER} system.

On a related note, there has been increased interest in improving the performance of \ac{SER} using \emph{personalisation}.
This approach acknowledges that the expression of emotion can be highly individualistic~\citep{Larsen87-AIA}.
Differences in expression may be influenced by culture~\citep{Russel94-ITU}, age~\citep{Gross97-EAA}, gender~\citep{Chaplin13-GDI}, or other factors, but also by temperamental differences across individuals~\citep{Larsen87-AIA, Sherman15-TIE}
Personalisation methods go beyond \emph{population-level} models (\ie models which are trained once and applied as-is to all new speakers during inference) and instead rely on \emph{speaker-level} models which typically fall under three categories:

\noindent
a) Methods which build speaker-level models; this can be done either by training/fine-tuning on data from a given individual, or training new models on subsets of the training data suited to that individual (\eg similar speakers)~\citep{Rudovic18-PML, Sridhar22-UPO, Song23-DMH, Kathan22-PDF}.
A typical example is that of \citet{Rudovic18-PML}, who introduce individual-level output heads; the data is first passed through a common backbone network trained for all speakers, and subsequently through output layers that are only trained with data from each speaker.
This approach is similar to federated learning, which makes local updates to models using speaker data while simultaneously keeping a global model which aggregates model updates from all speakers~\citep{McMahan17-CEL}.
Alternatively, a model can be retrained with data of the target speaker, or (additionally) with data from the most similar speakers to the target one during training~\citep{Sridhar22-UPO}.
The downside of these methods is that they require data from the target speaker to be already available during training, or on-the-fly retraining for each new speaker.

\noindent
b) Methods which introduce additional personal information, \eg in the form of demographic metadata (sex, age, etc.) that are given as extra inputs to the model~\citep{Hulsen19-FBD, Gerczuk23-ZSP}. 
This approach is typically pursued in the scope of \emph{precision medicine}~\citep{Kosorok19-PME}, which aims to provide personalised treatment by accounting for individual patient histories and characteristics.
While previous works have shown to improve performance with this approach on speech-based tasks as well~\citep{Gerczuk23-ZSP}, they do not capitalise on information about how a speaker actually sounds as they only exploit different `modalities', thus leaving a gap to be covered by the last family of methods.

\noindent
c) Methods which adapt population-level models in a \emph{few-shot} fashion; typically, those rely on a few enrolment samples~\citep{Triantafyllopoulos21-DSC, Triantafyllopoulos22-ESE, Fan22-IIS}.
The upside of the these approaches is that they require no re-training of the model during inference.
This allows for a streamlined deployment phase without any changes to the model -- a significant benefit given the ever-increasing complexity of contemporary \acp{DNN}.
Examples include \citet{Triantafyllopoulos21-DSC} and \citet{Fan22-IIS}, who use one or more neutral samples to condition an \ac{SER} network, or \citet{Triantafyllopoulos22-ESE} who use two randomly selected samples in an analogous way.
Similarly, \citet{Rahman12-APE} iteratively normalise features for each speaker; the added benefit of this method is that it does not require any label information, although more samples are required to obtained robust normalisation parameters and the method might not be suited to \acp{DNN}~\citep{Triantafyllopoulos21-DSC}.
We note that these methods are different from attempts to extract speaker-specific emotion predictions by disentangling emotional and speaker information~\citep{Le21-SAE, Yin20-SIA}; essentially, personalisation, as we define it, boils down to \emph{adaptation} to individual speaker characteristics, rather than \emph{invariance} to them.

In the present contribution, we expand on both research directions.
We introduce alternative considerations for \emph{individual fairness}, inspired by definitions of \emph{utility} and \emph{fairness} in economic theory.
Additionally, we revisit \emph{personalisation via enrolment} in an attempt to improve on those metrics.
Our methods are described in \cref{sec:methodology}, with results and discussion following in \cref{sec:results}.
We summarise in \cref{sec:conclusion}.

\section{Methodology}
\label{sec:methodology}

In this section, we begin with a description of the datasets we used in \cref{ssec:datasets}, followed by our personalisation method in \cref{ssec:method}, and experimental settings in \cref{ssec:settings}.
Our individual-level evaluation scheme is presented in \cref{ssec:fairness}.
\subsection{Datasets}
\label{ssec:datasets}

\textbf{FAU-AIBO} is a standard, categorical \ac{SER} dataset used in the INTERSPEECH 2009 Emotion Challenge~\citep{Schuller09-TI2a}. %BS: Please keep the original name - I changed.
The data is collected in a Wizard-of-Oz scenario where a remotely-controlled robot (`AIBO') interacts with children between the ages of 6 and 10.
The study participants attended one of two schools from the same region in Germany, (\emph{Ohm}) and (\emph{Mont}), with \emph{Ohm} being used as the training set and \emph{Mont} as the test set in the original challenge partition scheme.
As no validation set was defined, we use the last two speakers (\emph{Ohm\textsubscript{31}}, \emph{Ohm\textsubscript{32}}) of the training set, similar to \citep{Triantafyllopoulos21-DSC}.
The collected data has been annotated for $11$ classes by $5$ individual raters on the word-level.

For the challenge, the $11$ original classes were mapped to two alternative formulations of emotion:
a 2-class problem, where participants had to differentiate between \emph{negative} ($NEG$) and \emph{non-negative} ($IDL$) emotions;
and a 5-class problem, where participants had to classify an utterance as \emph{angry} ($A$), \emph{neutral} ($N$), \emph{motherese}/\emph{joyful} ($P$), \emph{emphatic} ($E$), with a 5\textsuperscript{th} \emph{rest} ($R$) class.
Additionally, the original words were manually aggregated to semantically and prosodically meaningful chunks, with a chunk label derived from the word-level labels using a heuristic process~\citep{Steidl09-ACO}.
The resulting data is highly imbalanced, and are dominated by neutral/non-negative states.
For this reason, we use the \ac{UAR} 
%BS: This is best described:
-- the added recall per class devided by the number classes -- 
to measure performance in the presence of class distribution imbalance, following the challenge specifications.
%BS: Why "similar"? I changed to following - please check.

\textbf{MSP-Podcast} is a recent, large-scale \ac{SER} dataset annotated both for categorical emotions and dimensional attributes~\citep{Lotfian17-BNE}.
The data has been annotated for 9 emotional classes, plus an extra neutral class and another one for instances where annotators disagree. % (\emph{``no agreement''}).
We use the latest version available at the time of submission (\textsc{v1.11.0}), and focus exclusively on the standard $4$-class problem pursued in most other recent \ac{SER} works~\citep{Triantafyllopoulos23-MLC}, with the set of labels being \{\emph{angry}, \emph{happy}, \emph{neutral}, \emph{sad}\} and exclude all other data.
We end up with $44\,586$ instances in the training set, $11\,947$ in the validation, and $20\,845$ in the test set.
We only use the more recent \textsc{Test1} partition and exclude \textsc{Test2}.

\subsection{Personalisation via enrolment}
\label{ssec:method}

\begin{figure}
    \centering
    \fbox{\includegraphics[width=\columnwidth]{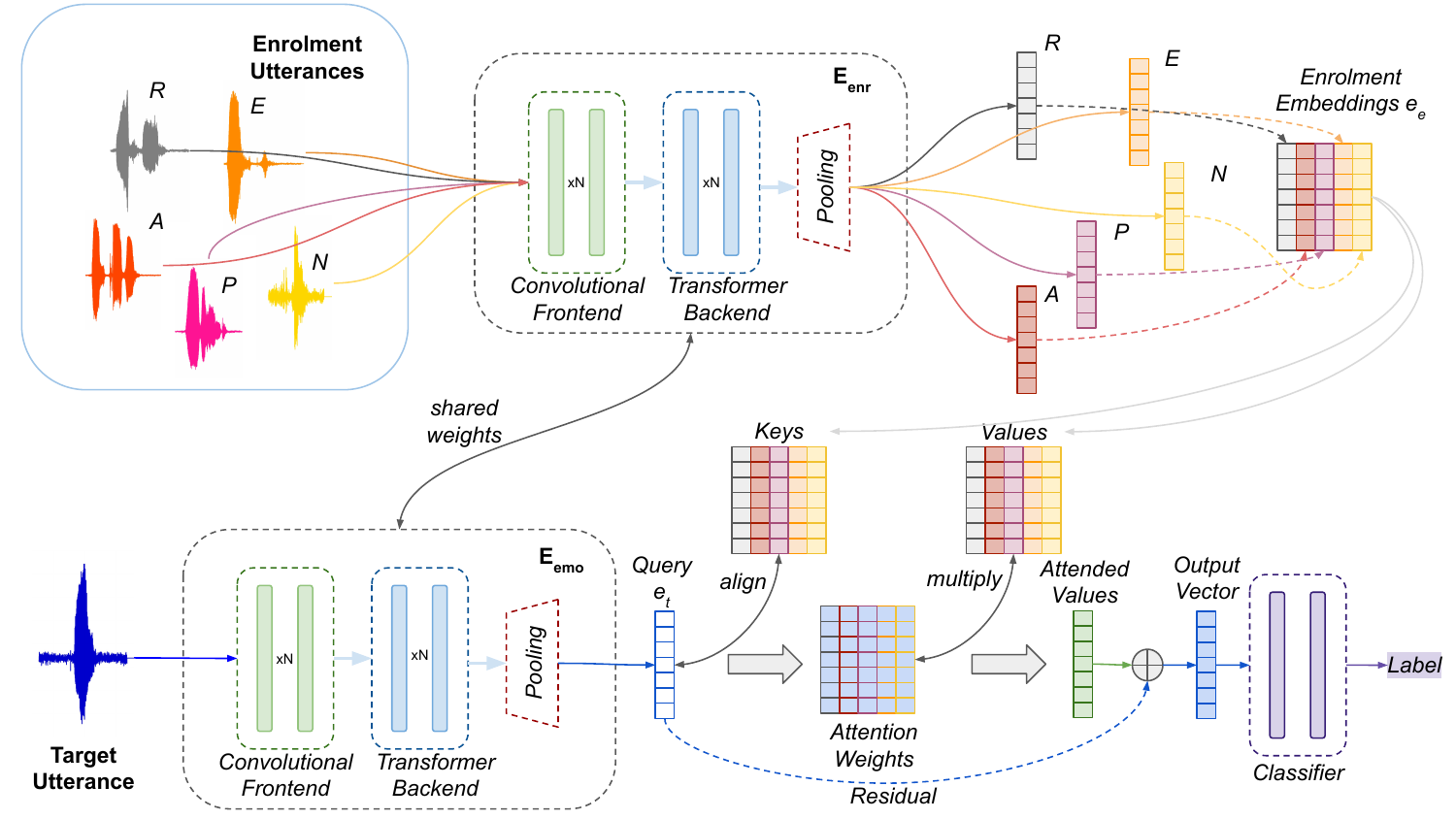}}
    \caption{
    Overview of the proposed architecture.
    A set of enrolment utterances is passed through an encoder to generate enrolment embeddings.
    These are used as keys and values in a dot-product attention scheme with the embeddings generated from the target utterance.
    The output embedding is passed to a feed-forward neural network for the final classification.
    Weights are shared between the enrolment and the main encoders.
    }
    \label{fig:arch}
\end{figure}

An overview of our architecture is shown in \cref{fig:arch}\footnote{Code: {\scriptsize \url{https://github.com/ATriantafyllopoulos/enrollment-personalization}}}.
Our workflow encapsulates the following key principles:

\noindent
a) Following standard \ac{DL} practice, the target utterance $u_t$ is passed through an emotion encoder $E_{emo}(\cdot)$ to generate a target embedding $e_t$.
$E_{emo}(\cdot)$ can be any \ac{DNN}; here, we opt for a \textsc{wav2vec2.0} model fine-tuned for dimensional \ac{SER}~\citep{Wagner23-DOT}, as we expect it to already have a good representation of emotion.
% \footnote{\url{https://huggingface.co/audeering/wav2vec2-large-robust-12-ft-emotion-msp-dim}}

\noindent
b) A set of \emph{enrolment} utterances $\{u_e\}$, coming from the same speaker as the target utterance, are used to \emph{adapt} an \ac{SER} system to new, previously unseen speakers.
In general, these $\{u_e\}$ can be exemplars of every potential label supported by the \ac{SER} model or a subset thereof.
In practice, we experiment with different alternatives as discussed below.

\noindent
c) These enrolment utterances are passed through an enrolment encoder $E_{enr}(\cdot)$ to generate suitable embeddings $\{e_e\}$.
$E_{enr}(\cdot)$ can either be different from the $E_{emo}(\cdot)$ used for $u_t$, be identical to it but trained separately, or even share the same weights.
We opt for the latter option as we intend the enrolment encoder to capture the emotional information present in $\{u_e\}$ and project it to the same embedding space as the target utterance.

\noindent
d) We then combine the enrolment embeddings $\{e_e\}$ with the target embedding $e_t$.
We choose a multihead, dot-product attention mechanism for this combination.
The enrolment embeddings are first concatenated and form a set of \emph{keys} ($K$), with the target embedding functioning as the \emph{query} ($Q$).
Following \citet{Vaswani17-AIA}, we first compute the outer product of the keys and queries ($QK^T$) and then pass it through a softmax function to generate a `soft' similarity estimate of the query with each key; for numerical stability the output is divided by $\sqrt{(d_k)}$, with $d$ being the dimensionality of the key embeddings.
This soft similarity matrix is subsequently multiplied with the \emph{values} ($V$) to produce the final output which is added to the target embedding to generate the final output (\ie the attention mechanism features a `residual' connection).
As values, we also use the embeddings of the enrolment utterances.

\noindent
e) The final product of the attention is passed to an \ac{MLP} classifier and the whole system is trained end-to-end.
% which produces the output label.
% The whole system is trained end-to-end.

Intuitively, this process of attention computes a (soft) similarity of the target utterance with each enrolment utterance, then combines the emotional information in these enrolment samples with the target sample to improve classification performance.

\subsection{Experimental settings}
\label{ssec:settings}

\noindent
\textbf{Enrolment utterances}\footnote{We include CSV files with the filename used for training/development/test set and the corresponding enrolment sets as supplementary material.}
Our approach requires setting aside a set of enrolment utterances for each speaker.
We always include a single enrolment utterance per class (though some variants of our approach do not make use of all of them; see below).
When a class is missing for a particular speaker, we impute it with zeros (\ie silence).
To create this enrolment set, we first sort the utterances of each speaker alphabetically (so in their order of appearance following the naming scheme of FAU-AIBO~\citep{Steidl09-ACO} and MSP-Podcast~\citep{Lotfian17-BNE}) and then for each class select the first utterance in which it appears.
% As for FAU-AIBO, we keep as enrolment utterances those utterances where each emotion appears for the first time (alphabetically) for each speaker and impute missing classes with zeros.

\noindent
\textbf{Personalisation setup:}
In total, we investigate three alternative formulations of the personalisation problem and contrast their effectiveness with respect to a baseline model:
\begin{enumerate}
    \item \base -- Our baseline model only includes the emotion encoder and the downstream \ac{MLP} classifier; note that this setup is identical to recent state-of-the-art work for dimensional \ac{SER}~\citep{Wagner23-DOT} and we thus expect it to be a strong baseline.
    \item \persn -- For our first personalisation approach, we only utilise neutral utterances for the enrolment; this is most similar to the setup of \citet{Triantafyllopoulos21-DSC}, who condition their model on a single neutral utterance from each speaker.
    \item \perse -- We also experiment with using only the non-neutral (\ie emotional or rest) utterances for enrolment.
    \item \persa -- Finally, we use all available enrolment utterances (both neutral and emotional).
\end{enumerate}

\noindent
\textbf{Hyperparameters:}
We train all models for $50$ epochs with an Adam optimiser, a learning rate of $0.0001$, and a batch size of $4$, all standard hyperparameters from previous literature~\citep{Wagner23-DOT}.
We select the epoch with the best \ac{UAR} on the validation set for our final evaluation on the test set.

\subsection{Individual-level fairness}
\label{ssec:fairness}

The standard process for computing performance is to consider each chunk as an independent trial.
This process was also adopted in the 2009 INTERSPEECH Emotion Challenge, and we denote its outcome as $UAR_{C}$.

To define individual-level fairness, we begin by computing the performance on a speaker-level, thus assuming that first speakers are selected independently, and only then are samples selected independently for them (see \citep{Guyon98-WST} for a similar argumentation).
We finally compute the \ac{UAR} over the set of chunks for each individual speaker in our dataset, which we denote as $UAR_{SP}$.
We call this the \textbf{utility} of each speaker, as this is the benefit that each speaker can expect from getting their emotions recognised correctly\footnote{Naturally, what this `benefit' entails depends entirely on the downstream application.}.
We examine this utility under three different perspectives:

\noindent
\textbf{I) Statistics:}
We first report standard statistics, such as the mean ($\mu(\cdot)$), median ($\mu_{1/2}(\cdot)$), standard deviation ($\sigma(\cdot)$), maximum ($\max(\cdot)$), and minimum ($\min(\cdot)$) of $UAR_{SP}$.
These statistics give us a coarse characterisation of how utility is distributed across the different individuals in our dataset.

\noindent
\textbf{II) Gini coefficient:} The Gini coefficient ($G(\cdot)$) is a standard measurement used in econometrics to judge the distribution of utility in a particular population and is defined as half of the mean absolute difference relative to the mean of a particular sample~\citep{Dorfman79-AFF}.
In particular, it takes the value of $0$ for an equal distribution where everyone has the same utility, and $1$ for a completely unequal one, where the entire utility is accumulated by one particular individual.
In our case, this is computed as:
\begin{equation}
    G(\{u_i\}_1^N) = \frac{\sum\limits_{i=1}^{i=N}\left(\sum\limits_{j=1}^{j=N}(|u_i-u_j|)\right)}{\mu(\{u_i\}_1^N)},
\end{equation}
where $N$ is the number of speakers in the test set and $\{u_i\}_1^N$ is the set of utility values for all speakers, with $u_i = UAR_{SP}^i$, \ie the speaker-level \ac{UAR} computed for each speaker.

\noindent
\textbf{III) \Acp{ISWF}:} Besides the Gini coefficient, the distribution of utility can also be considered under alternative formulations.
A classical formalisation of the problem is that given by \citet{Atkinson70-OTM}, who defines a family of \acp{ISWF}:
\begin{equation}
    \label{eq:iswf}
    W_{\alpha}(u_1, ..., u_N) =
    \begin{cases}
      & \frac{1}{N}(\sum\limits_{i=1}^{i=N}{u_i^{1-\alpha}})^{\frac{1}{1-\alpha}} \quad \text{if} \ \alpha \neq 1
      \\[10pt]
      & \sqrt[N]{\prod\limits_{i=1}^{i=N}u_i} \quad \text{if} \ \alpha = 1 \text{,}
    \end{cases}.
\end{equation}
These \acp{ISWF} allow for a more modular definition of fairness, as they do not presuppose an equal distribution of utility as the most fair outcome (like the Gini coefficient).
For example, as $\alpha \rightarrow \infty$, \cref{eq:iswf} approximates Rawls' ``difference principle'' (\ie maximise the minimum utility)~\citep[Ch. II, § 13, pg. 65]{Rawls99-ATO}, whereas as $\alpha \rightarrow 0$, it approaches the standard utilitarian approach of maximising total utility irrespective of fairness.
Thus, \acp{ISWF} provide a modular `knob' that allows stakeholders to define fairness for their particular needs.
In our work, we compute the value of the \ac{ISWF} for different values of $\alpha \in \{0, 100\}$, measuring the suitability of different models for different scenarios.

\section{Results \& Discussion}
\label{sec:results}

\begin{table*}[t]
    \centering
    \caption{
    Global and individual-level performance for the $2$- and $5$-class problems of FAU-AIBO, as well as the $4$-class problem of MSP-Podcast.
    We compute the $UAR_C$ over all chunks in the test set, along with $95\%$ confidence intervals (CIs) obtained via bootstrapping.
    Furthermore, we compute fairness metrics on speaker-level UAR ($UAR_{SP}$).
    For easier comparison, we mark metrics as ascending ($\uparrow$, higher is better) and descending ($\downarrow$, lower is better).
    }
    \label{tab:results}
\resizebox{\textwidth}{!}{
    \begin{tabular}{c|c|cccccc}
        \toprule
        \textbf{Method} & $UAR_C (\uparrow)$ & $G(UAR_{SP}) (\downarrow)$ & $\mu(UAR_{SP})(\uparrow)$ & $\sigma(UAR_{SP}) (\downarrow)$ & $\mu_{1/2}(UAR_{SP})(\uparrow)$ & $max(UAR_{SP}) (\uparrow)$ & $min(UAR_{SP}) (\uparrow)$\\
        \midrule
        \multicolumn{8}{c}{\textbf{FAU-AIBO (2-class problem)}}\\
        \midrule
        \base & .677 [.666 - .689] & .091 & .699 & .115 & .686 & 1.000 & .500\\
        \persn & .680 [.669 - .692] & .097 & .697 & .120 & .715 & 1.000 & .500\\
        \perse & .700 [.690 - .711] & .082 & .703 & .107 & .731 & 1.000 & .500\\
        \persa & \textbf{.718 [.707 - .729]} & \textbf{.073} & \textbf{.728} & \textbf{.098} & \textbf{.735} & 1.000 & .500\\
        \midrule
        \multicolumn{8}{c}{\textbf{FAU-AIBO (5-class problem)}}\\
        \midrule
        \base & .443 [.426 - .461] & .160 & .406 & .132 & .417 & .806 & .053\\
        \persn & .424 [.408 - .442] & .118 & .399 & .086 & .390 & .614 & \textbf{.214}\\
        \perse & .440 [.422 - .460] & .137 & .432 & .137 & \textbf{.419} & \textbf{1.000} & .207\\
        \persa & \textbf{.452 [.434 - .469]} & \textbf{.106} & \textbf{.428} & \textbf{.084} & .412 & .576 & .189\\
        \midrule
        \multicolumn{8}{c}{\textbf{MSP-Podcast (4-class problem)}}\\
        \midrule
        \base & \textbf{.567 [.559 - .575]} & .242 & .438 & .200 & .413 & 1.000 & .000\\
        \persn & .510 [.502 - .520] & .233 & .361 & \textbf{.170} & .333 & 1.000 & .000\\
        \perse & .563 [.554 - .573] & .215 & .460 & .184 & .436 & 1.000 & .000\\
        \persa & .563 [.554 - .572] & \textbf{.201} & \textbf{.479} & .186 & \textbf{.442} & 1.000 & .000\\
        \bottomrule
    \end{tabular}
}
\end{table*}

% \begin{figure}[t]
%     \centering
%     \includegraphics[width=.49\columnwidth]{figs/baseline.pdf}~%
%     % \includegraphics[width=.24\textwidth]{figs/neutral.pdf}~%
%     % \includegraphics[width=.24\textwidth]{figs/emotional.pdf}~%
%     \includegraphics[width=.49\columnwidth]{figs/all.pdf}
%     \caption{
%     Confusion matrices for the baseline and our personalisation model using enrollment from all classes.
%     }
%     \label{fig:confusion}
% \end{figure}

\begin{figure}[t]
    \centering
    \includegraphics[width=\columnwidth]{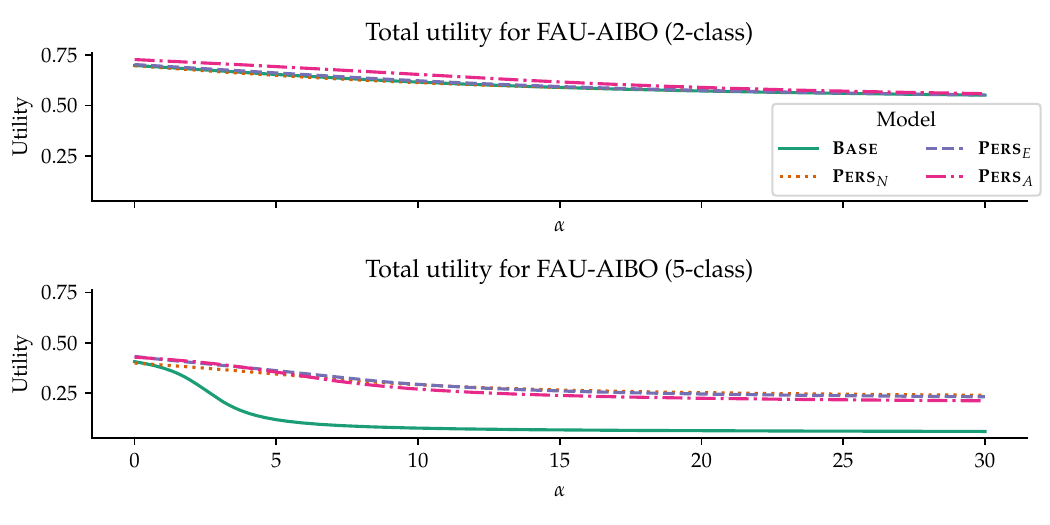}
    \caption{
    Total utility achieved by each model for different isoelastic social welfare functions for the $2$- (top) and $5$-class (bottom) formulations of FAU-AIBO.
    Utility is defined as $UAR_{SP}$.
    }
    \label{fig:iswf}
\end{figure}

\cref{tab:results} shows our results.
It includes the global \ac{UAR} ($UAR_{C}$), which is computed over all chunks in the data (including 95\% CIs), as well as the different fairness statistics and the Gini index for both the $2$- and the $5$-class problem.
\persa shows the best performance overall, with higher $UAR_{C}$ than all alternatives for FAU-AIBO and marginally lower than the baseline for MSP-Podcast, as well as overall better performance with respect to different individual-level metrics.
Specifically, it improves from a $67.7\%$ to a $71.8\%$ $UAR_{C}$ for the $2$-class problem, and from $44.3\%$ to $45.2\%$ for the $5$-class problem for FAU-AIBO.
More importantly, it improves across all fairness metrics for all datasets and tasks, showing a lower Gini index and standard deviation, as well as a higher mean and median for speaker-level \ac{UAR} ($UAR_{SP}$) -- our measure of utility.
It is only surpassed by \persn on the standard deviation of performance for MSP-Podcast, where it is still outperforming the baseline model.
Generally, the same can be said for \persn and \perse, as both show better performance than the baseline, with the latter additionally outperforming the former.

% In addition, \cref{fig:confusion} shows the confusion matrices for the baseline and \textbf{Pers (A)} on the $5$-class problem.
% We observe that most of the gains seem to stem from better recognising the \emph{emphatic} class ($E$), with all classes except \emph{rest} ($R$) generally benefiting.
% This shows how the addition of enrolment personalisation changes the overall behaviour of a model, though the benefits may not always be consistent.
% Nevertheless, we note that the \emph{rest} class is highly ambiguous for FAU-AIBO, as it encompasses multiple classes and an overall lower level of human agreement.

\cref{fig:iswf} additionally shows how the total utility amassed by each model changes for different \acp{ISWF}.
Due to space limitations, we only show results for FAU-AIBO.
For the $2$-class problem, \persa results in higher utility for most values of $\alpha$, while converging to the other methods as $\alpha \rightarrow \infty$.
For the $5$-class problem, we observe that \base quickly approaches zero utility according to the difference principle ($\alpha \rightarrow \infty$), with the three personalisation models showing similar behaviour.
% These curves illustrate how the models might behave different with respect to the different fairness definitions.
For all models, utility is highest for $\alpha \rightarrow 0$, \ie the utilitarian scenario, where average utility is maximised without consideration for its distribution.
However, the total utility quickly drops as $\alpha$ increases, reflecting scenarios where the discrepancy between different speakers becomes more important.

Broadly, the decision on which the $\alpha$ value or fairness metric is appropriate for a particular application rests with the stakeholders that are affected by it.
For example, recent work employed an `equality of outcome' requirement for different users across multiple \ac{ML} algorithms (in the context of recommendation algorithms)~\citep{Sharifi19-AIF}.
This would be equivalent to a Gini index of 0 in our definition.
Digital health applications, on the other hand, may require a maximisation of a lower bound on speaker-level performance -- in this case, a larger $\alpha$ would be more appropriate, to place more emphasis on the worst-performing speakers~\citep{Triantafyllopoulos23-H4H}.

Collectively, our results show that personalisation via enrolment can improve predictive performance and make this performance more equal across different speakers.
This is vital for providing a uniform and fair user experience in applications relying on \ac{SER}.
Another interesting finding is that models personalised on all classes or even only all emotional classes generally outperform those personalised only on neutral data.
This is in contrast to the prior work which has used these neutral enrolment utterances for personalisation~\citep{Triantafyllopoulos21-DSC, Fan22-IIS}.

\section{Conclusion}
\label{sec:conclusion}

We have introduced a novel method for personalisation using a minimal set of enrolment utterances -- one per class.
Our method relies on dot-product attention for injecting information from these utterances into a classification network.
Additionally, we introduced novel considerations for individual-level fairness that takes into account performance on the individual level.
Overall, we showed how our method can improve performance both on the global level and for the fairness metrics we introduced.
\textbf{Limitations:}
% Due to space limitations, we only have benchmarked our method on a single dataset (albeit with two tasks).
% To ensure that the method truly generalises, it is necessary to test it on other data as well.
Our method depends on providing accurate enrolment samples, and may fail if (intentionally) given erroneous ones.
\textbf{Future work:} %BS: slightly shortened for space.
Alternative methods for introducing the enrolment information %into an \ac{SER} model 
can be investigated.
Furthermore, explainability methods can be used to understand how the network is using the additional enrolment information, \eg by visualising and probing the embedding space before and after the enrolment information is injected.
Finally, ways for safeguarding against inaccurate enrolment samples must be developed. % to improve the robustness of the method.
% \lipsum[50]

% \newpage
%BS: are we allowed to have this on page 5? We can insert later on, anyhow, so left out for now.
\section{Acknowledgements}
This work was partially funded by the EU H2020 project No.\ 101135556 (INDUX-R).

%BS: Please try to fill with references until page end leaving only space for acknowledgement
\section{\refname}
\printbibliography[heading=none]

\end{document}